\documentclass{article}
\usepackage{spconf,amsmath,graphicx,hyperref}
\usepackage{threeparttable} 
\usepackage{booktabs}
\usepackage{multirow}
\usepackage{tabularx}
\usepackage{amssymb} 
\usepackage{hyperref} 
\usepackage{cite}

\title{ULW-SleepNet: An Ultra-Lightweight Network for Multimodal Sleep Stage Scoring}
\name{
Zhaowen Wang$^{1,2}$,
Dongdong Zhou$^{1,3,\ast}$,
Qi Xu$^{1,3}$, Fengyu Cong$^{4,3,2}$,
Mohammad Al-Sa’d$^{\,5,6}$,
Jenni Raitoharju$^{2}$
\thanks{$\ast$ Corresponding author: zhoudd@dlut.edu.cn.}
}

\address{
$^{1}$ School of Computer Science and Technology, Dalian University of Technology, Dalian, China \\
$^{2}$ Faculty of Information Technology, University of Jyväskylä, Jyväskylä, Finland \\
$^{3}$ Key Laboratory of Social Computing and Cognitive Intelligence, Ministry of Education, Dalian, China \\
$^{4}$ School of Biomedical Engineering, Faculty of Medicine, Dalian University of Technology, Dalian, China \\
$^{5}$ Department of Computing Sciences, Tampere University, Tampere, Finland \\
$^{6}$ Faculty of Medicine, University of Helsinki, Helsinki, Finland
}

\begin{document}
\ninept
\maketitle
\begin{abstract}
Automatic sleep stage scoring is crucial for the diagnosis and treatment of sleep disorders. Although deep learning models have advanced the field, many existing models are computationally demanding and designed for single-channel electroencephalography (EEG), limiting their practicality for multimodal polysomnography (PSG) data. To overcome this, we propose ULW-SleepNet, an ultra-lightweight multimodal sleep stage scoring framework that efficiently integrates information from multiple physiological signals. ULW-SleepNet incorporates a novel Dual-Stream Separable Convolution (DSSC) Block, depthwise separable convolutions, channel-wise parameter sharing, and global average pooling to reduce computational overhead while maintaining competitive accuracy. Evaluated on the Sleep-EDF-20 and Sleep-EDF-78 datasets, ULW-SleepNet achieves accuracies of 86.9\% and 81.4\%, respectively, with only 13.3K parameters and 7.89M FLOPs. Compared to state-of-the-art methods, our model reduces parameters by up to 98.6\% with only marginal performance loss, demonstrating its strong potential for real-time sleep monitoring on wearable and IoT devices. The source code for this study is publicly available at https://github.com/wzw999/ULW-SLEEPNET.
\end{abstract}
\section{Introduction}
\label{sec:intro}

Sleep is a vital physiological process essential for human health, yet the prevalence of sleep disorders is steadily increasing \cite{berry2017aasm}.  Accurate sleep stage scoring is therefore critical for diagnosing and treating conditions such as insomnia and sleep apnea \cite{kales1970evaluation}. This scoring process is commonly performed in accordance with established standards, such as those provided by the American Academy of Sleep Medicine (AASM) \cite{berry2017aasm}. The AASM guidelines define five sleep stages: Wake, three Non-Rapid Eye Movement (NREM) stages (N1, N2, and N3), and the Rapid Eye Movement (REM) stage. Each stage is defined by unique patterns in physiological signals. Polysomnography (PSG), the clinical gold standard, assesses sleep architecture through simultaneous recording of multiple signals, including electroencephalography (EEG), electrooculography (EOG), and electromyography (EMG). However, the reliance on manual scoring by sleep experts is not only time-consuming but also prone to inter-scorer variability due to the subjective interpretation of scoring rules \cite{zhou2024interpretable}.

The limitations of traditional manual scoring have motivated the widespread exploration of automatic deep learning-based methods \cite{nasiri2025caisr, xu2022convolutional, jia2024distillsleepnet, yan2021automatic, zhao2025selectivefinetuning}. Nevertheless, many existing models suffer from high computational complexity, limiting their practicality for wearable devices and real-time applications. This necessitates the development of lightweight architectures that can deliver both high accuracy and efficiency under practical constraints. In response, several recent studies have focused on designing computationally efficient models \cite{zhou2022singlechannelnet,   lv2024ssleepnet, ren2025flexiblesleepnet, goerttler2025msa}. For instance, Fiorillo \emph{et al}. proposed a streamlined lightweight network that processes 90-second EEG segments and employs Monte Carlo dropout for uncertainty quantification \cite{fiorillo2021deepsleepnet}. Zhou \emph{et al}. utilized spectro-temporal representations of EEG signals to lower model complexity and expedite training \cite{zhou2021lightsleepnet}. Yang \emph{et al}. and Wang \emph{et al}. incorporated attention mechanisms and depthwise separable convolutions to enhance feature extraction from sleep data \cite{yang2023lwsleepnet, wang2025efficientsleepnet}. Liu \emph{et al}. proposed a novel framework integrating sub-Nyquist sampling and spiking neural networks to achieve ultra-low-power operation \cite{liu2025picosleepnet}. Despite these advances, most methods are tailored primarily to single-channel EEG, underscoring a continuing demand for even more efficient architectures capable of handling multimodal sleep data.

\begin{figure*}
   \centering
   \includegraphics[width=1\textwidth]{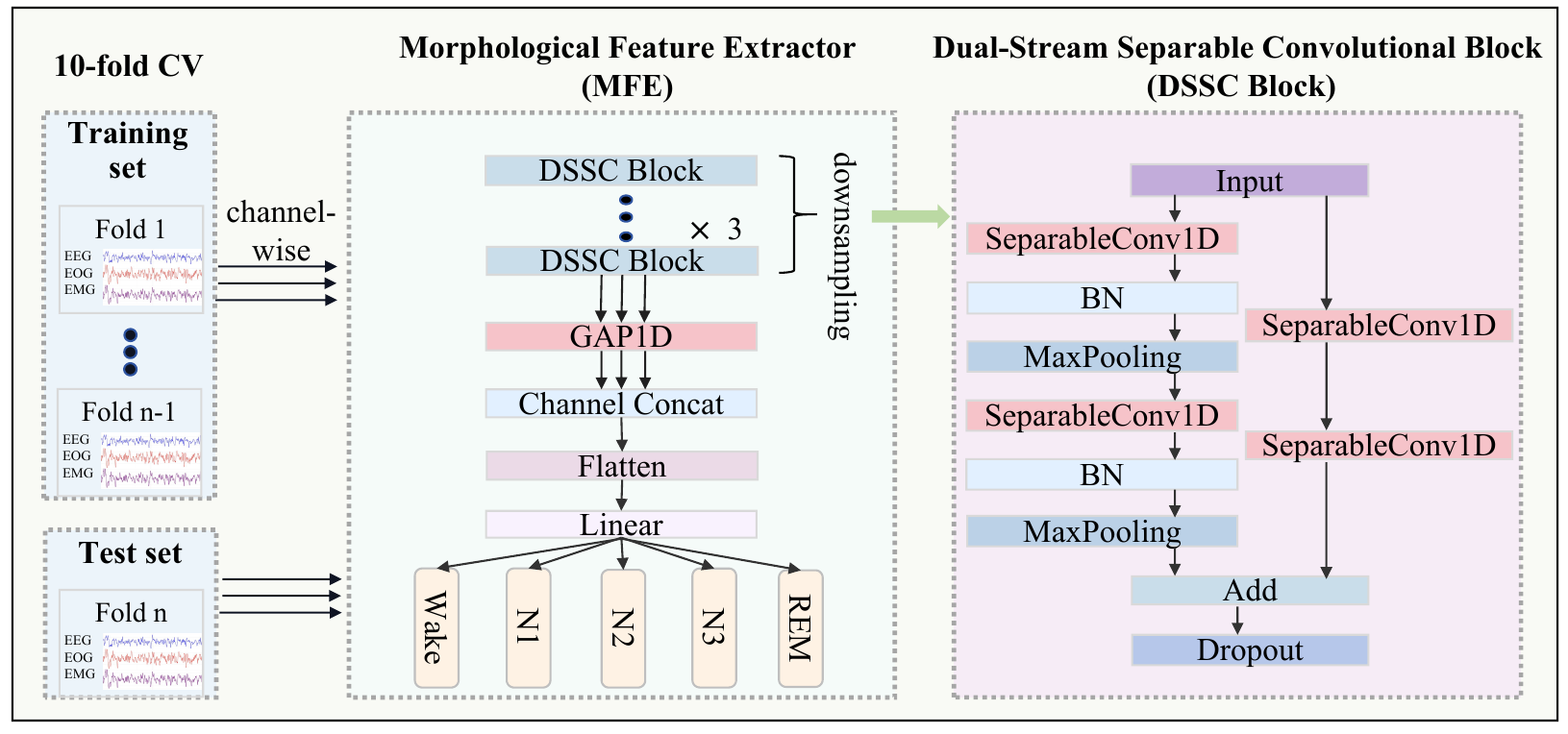}
    \caption{An overview of the ULW-SleepNet architecture. Multimodal physiological input signals (i.e., EEG, EOG, EMG) are processed through a shared channel-wise feature extraction pipeline based on the Dual-Stream Separable Convolutional Block.}
    \label{fig:structure}
\end{figure*}   

To address the aforementioned challenges, we propose ULW-SleepNet, whose architecture efficiently integrates information from multiple physiological signal channels, producing more robust and reliable sleep stage predictions. Our approach offers three main contributions:


1) We propose ULW-SleepNet, an ultra-lightweight and efficient sleep stage scoring framework especially for multimodal data, which efficiently scales with the number of input channels while maintaining low computational overhead.


2) We design a novel Dual-Stream Separable Convolution (DSSC) Block that facilitates efficient residual learning and incorporates key lightweight design strategies, enabling the capture of both transient events and long-term temporal features to enhance sleep stage scoring performance while maintaining an ultra-lightweight architecture.

3) We evaluate ULW-SleepNet on two public PSG datasets, Sleep-EDF-20 and Sleep-EDF-78, demonstrating a superior balance between performance and efficiency by achieving high accuracy with significantly reduced parameters and computational cost suitable for practical deployment.


\section{Methodology}
\label{sec:motho}

\subsection{ULW-SleepNet Architecture Overview}

ULW-SleepNet is an ultra-lightweight framework specifically optimized for multimodal sleep stage scoring with low computational overhead. As illustrated in Fig. \ref{fig:structure}, the model incorporates a morphological feature extractor that efficiently processes multi-channel physiological signals, with the DSSC Block serving as its core component.

\subsection{Key Lightweight Design Strategies}


Designing ULW-SleepNet as an ultra-lightweight architecture necessitates a balance between computational efficiency and the ability to extract discriminative features from multimodal physiological signals. To this end, the network employs a set of complementary strategies aimed at reducing both parameter count and computational complexity while maintaining robust performance. Below, we elaborate on four core design strategies central to ULW-SleepNet: the use of depthwise separable convolutions, parameter-shared channel-wise processing, the DSSC Block architecture, and global average pooling for dimensionality reduction.

\subsubsection{Depthwise Separable Convolution Integration}
To achieve significant parameter reduction, we replace standard convolutions with depthwise separable convolutions throughout the network \cite{chollet2017xception}. A standard convolution operation requires $D_k \times M \times N$ parameters, where $D_k$ is the kernel size, $M$ is the number of input channels, and $N$ is the number of output channels. Depthwise separable convolution factorizes this into depthwise and pointwise operations, reducing computational complexity from $O(D_k \times M \times N)$ to $O(D_k \times M + M \times N)$.  With a kernel size of 3 ($D_k=3$), this approach reduces computational complexity to a ratio of approximately $1/N + 1/D_k$, significantly compressing the model while keeping key features.

\subsubsection{Channel-Wise Processing with Parameter Sharing} 

For multimodal sleep data with $C$ channels and $T$ time points, the input tensor is denoted as $X \in \mathbb{R}^{T \times C}$. Each channel is processed independently through the same lightweight feature extraction pipeline: $F_c = \phi(X_c),\quad c = 1, 2, ..., C$, where $\phi$ represents the shared feature extractor composed of the DSSC Blocks and a global average pooling layer, and $ X_c\in\mathbb{R}^{T \times 1}$ denotes the c-th channel's time series data with T time points. By sharing the same parameters across channels, the model reduces its size by a factor of $C$ compared to learning separate parameters for each channel, while maintaining constant complexity and still capturing channel-specific temporal patterns.

\begin{table*}[t]
\centering
\footnotesize
\caption{Ablation study on the Sleep-EDF-20 dataset.}
\label{tab:ablation}
\renewcommand{\arraystretch}{1.078}
\begin{tabularx}{\textwidth}{lXccc}
\toprule
\textbf{Ablated Component} & \textbf{Configuration} & \textbf{ACC (\%)} & \textbf{Params} & \textbf{FLOPs} \\
\midrule
\textbf{Proposed (ULW-SleepNet)} 
 & 3 blocks, ks = 3, ps = 2, fs = (8, 16, 32), SeparableConv1D, EEG/EOG/EMG & 86.88 & 13,337 & 7.89 M \\
\midrule
\textbf{Input modality} 
 & only EEG &  85.58 & 13,337 & 7.89 M \\
\midrule
\multirow{2}{*}{\textbf{DSSC Block Number}} 
 & 2 blocks, fs = (16, 32) & 85.37 & 12,841 & 17.39 M \\
 & 4 blocks, fs = (8, 16, 32, 64) & 86.32 & 34,713 & 10.39 M \\
\midrule
\textbf{Kernel Size \& Pooling Stride} 
 & ks = 7, ps = 4 & 86.57 & 13,661 & 9.19 M \\
\midrule
\multirow{3}{*}{\textbf{Filter Number}} 
 & fs = (4, 8, 16) & 84.64 & 5,873 & 2.54 M \\
 & fs = (16, 32, 64) & 86.49 & 34,217 & 27.22 M \\
 & fs = (16, 32, 64, 128) & 87.16 & 101,545 & 36.92 M \\
\midrule
\textbf{Conv Type}
 & Standard Conv1D & 86.32 & 16,997 & 15.03 M \\
\bottomrule
\end{tabularx}
\end{table*}

\subsubsection{DSSC Block Architecture} 

The core innovation of our framework lies in the DSSC Block, which implements a highly efficient form of residual learning. Automated sleep scoring requires the identification of both transient, short-term events (e.g., sleep spindles) and long-term waveform characteristics (e.g., delta waves). To address this, the DSSC Block is specifically designed to capture complex temporal dependencies through its main processing stream, while preserving the original signal's fundamental features via a residual shortcut. This dual-stream approach ensures that the model can learn hierarchical features without losing crucial low-level information.

The output of the block, is the element-wise sum of the main stream and the lightweight shortcut stream. The main stream is designed to capture temporal patterns through two sequential lightweight convolutional operations. Specifically, the input first passes through a 1D depthwise separable convolution, Batch Normalization (BN), a ReLU activation, and a MaxPooling operation. The output of this sequence is then processed by a second, identical series of operations. This stacked design progressively reduces the spatial dimensions of the input while extracting increasingly abstract features.

To enable the residual connection, the shortcut stream must perform equivalent downsampling. Unlike traditional residual blocks that use standard convolutions for dimension matching, our shortcut connection utilizes two sequential separable convolutions, each with a stride of 2. The progressive channel expansion across stacked DSSC Blocks (e.g., 8, 16, 32 channels) further balances representational capacity with computational cost, while the pervasive use of batch normalization ensures stable training and faster convergence.

\subsubsection{Global Average Pooling for Dimension Reduction} 


Instead of computationally expensive fully connected layers, we employ global average pooling for spatial dimension reduction by taking the average of all temporal steps for each feature map. This reduces the spatial dimensions to a single value per feature map, eliminating the need for large dense layers and reducing parameters by approximately 90\% compared to traditional flatten-and-dense approaches while providing implicit regularization \cite{lin2013network}.

\subsection{Model Configuration}


Through empirical analysis, the optimal lightweight configuration employs three cascaded DSSC Blocks with progressive channel expansion of (8, 16, 32), kernel size of 3, and pooling stride of 2. The channel-wise features are concatenated and processed through a classification head consisting of a single hidden layer with 64 units followed by ReLU activation, then mapped to the 5 sleep stages through a final linear layer with softmax activation for probability distribution output.


For model optimization, L2 regularization with $\lambda$ = 0.001 is applied to all separable convolution layers to encourage weight sparsity. Strategic dropout is employed with rates of 0.1 between DSSC Blocks and 0.3 in the classification head to prevent overfitting while maintaining low computational overhead. These lightweight design strategies collectively enable ULW-SleepNet to achieve efficient performance while maintaining competitive accuracy for practical sleep monitoring applications.

\section{Experiments and Results}
\label{sec:exp}

\begin{table*}[t]
\centering
\begin{threeparttable}
\footnotesize
\caption{Comparison with SOTA models on two datasets.}
\label{tab:results}
\renewcommand{\arraystretch}{1.078}
\begin{tabularx}{\textwidth}{c|c|ccX|XXXXX|cX}
\toprule
\multirow{3}{*}{Dataset} & \multirow{3}{*}{Model}
& \multicolumn{3}{c|}{Overall Performance}
& \multicolumn{5}{c|}{Per-stage F1 Score (\%)}
& \multicolumn{2}{c}{Resource} \\
\cmidrule(lr){3-5} \cmidrule(lr){6-10} \cmidrule(lr){11-12}
& & ACC (\%) & MF1 (\%) & $\kappa$ & Wake & N1 & N2 & N3 & REM & Params & FLOPs \\
\midrule
\multirow{9}{*}{Sleep-EDF-20}
& DeepSleepNet\cite{supratak2017deepsleepnet} & 82.0 & 76.9 & 0.76 & 84.7 & 46.6 & 85.9 & 84.8 & 82.4 & 24.7M & -- \\
& DeepSleepNet-lite\cite{fiorillo2021deepsleepnet} & 84.0 & 78.0 & 0.78 & 87.1 & 44.4 & 87.9 & 88.2 & 82.4& 600K & -- \\
& TinySleepNet\cite{supratak2020tinysleepnet} & 85.4 & 80.5 & 0.80 & 90.1 & \textbf{51.4} & 88.5 & 88.3 & 84.3 & 1.3M & -- \\
& AttnSleepNet\cite{eldele2021attention} & 84.4 & 78.1 & 0.79 & 89.7 & 42.6 & 88.8 & \textbf{90.2} & 79.0 & 520K & 60.9M \\
& LWSleepNet\cite{yang2023lwsleepnet} & 86.6 & 79.2 & 0.81 & 92.4 & 41.3 & \textbf{90.2} & 88.4 & 84.0 & 180K & 55.3M \\
& EfficientSleepNet\cite{wang2025efficientsleepnet} & 84.4 & 78.1 & 0.79 & 87.7 & 41.9 & 89.2 & 89.9 & 82.1 & 83.8K & -- \\
& PicoSleepNet\cite{liu2025picosleepnet} & 83.5 & 75.2 & -- & -- & -- & -- & -- & -- & 22.2K & \textbf{0.68M} \\
& \textbf{ULW-SleepNet} & \textbf{86.9} & \textbf{80.7} & \textbf{0.82} & \textbf{93.5} & 45.8 & 89.3 & 90.1 & \textbf{84.9} & \textbf{13.3K} & 7.89M \\

\midrule
\multirow{9}{*}{Sleep-EDF-78}
& DeepSleepNet\cite{supratak2017deepsleepnet} & 76.9 & 70.7 & 0.69 & 90.8 & 44.8 & 78.5 & 67.9 & 71.3 & 24.7M & -- \\
& DeepSleepNet-lite\cite{fiorillo2021deepsleepnet} & 80.3 & 75.2 & 0.73 & 91.5 & 46.0 & 82.9 & 79.2 & 76.4& 600K & -- \\
& TinySleepNet\cite{supratak2020tinysleepnet} & \textbf{83.1} & \textbf{78.1} & \textbf{0.77} & 92.8 & \textbf{51.0} & 85.3 & 81.1 & \textbf{80.3} & 1.3M & -- \\
& AttnSleepNet\cite{eldele2021attention} & 81.3 & 75.1 & 0.74 & 92.0 & 42.0 & 85.0 & 82.1 & 74.2 & 520K & 60.9M \\
& LWSleepNet\cite{yang2023lwsleepnet} & 81.5 & 74.3 & 0.75 & 92.2 & 40.2 & \textbf{86.9} & 78.4 & 73.8 & 180K & 55.3M \\
& EfficientSleepNet\cite{wang2025efficientsleepnet} & 80.8 & 75.3 & 0.74 & 92.2 & 43.2 & 84.4 & \textbf{82.4} & 74.3 & 83.8K & -- \\
& PicoSleepNet\cite{liu2025picosleepnet} & 77.9 & 68.1 & -- & -- & -- & -- & -- & -- & 25.8K & \textbf{0.73M} \\
& \textbf{ULW-SleepNet} & 81.4 & 74.0 & 0.74 & \textbf{93.0} & 39.3 & 83.8 & 79.6 & 74.4 & \textbf{13.3K} & 7.89M \\
\bottomrule
\end{tabularx}
\begin{tablenotes}
\item Results of other methods are directly quoted from the corresponding original papers.
\end{tablenotes}
\end{threeparttable}
\end{table*}

\subsection{Datasets}
We evaluate our model on two publicly released subsets of the Sleep-EDF Expanded database, namely Sleep-EDF-20 and Sleep-EDF-78 \cite{kemp2000analysis}.

1) Sleep-EDF-20: It contains 39 PSG recordings from 20 healthy, medication-free subjects (10 male, 10 female, aged 25-34). For our analysis, we utilize dual-channel EEG (Fpz–Cz and Pz–Oz), a horizontal EOG channel, and a submental EMG channel, all sampled at 100 Hz. Following preprocessing steps outlined in \cite{zhou2022alleviating}, we exclude extended periods of wakefulness (Stage Wake) at the beginning and end of the recordings, retaining only the 30-minute segments immediately preceding and following the main sleep period, and discard any epochs labeled 'MOVEMENT' or 'UNKNOWN'. 
In addition, the original EEG signals are band-pass filtered with cutoff frequencies of 0.3 Hz and 45 Hz.

2) Sleep-EDF-78: It is a larger collection that includes 197 PSG recordings from 78 healthy subjects (40 male, 38 female, aged 25-101). The same signal configuration and preprocessing pipeline as in the Sleep-EDF-20 dataset are applied.

\subsection{Experimental setting}


The model is implemented in TensorFlow and trained using the Adam optimizer with a batch size of 32. The initial learning rate is set to $10^{-3}$ and dynamically adjusted using cosine annealing scheduling to ensure stable convergence over 50 training epochs. To ensure robust performance evaluation, we adopt a 10-fold cross-validation (CV) scheme with subject-wise data splitting, preventing data leakage across subjects and providing reliable generalization assessment.

\subsection{Ablation study}
To validate the architectural design and assess the contribution of each component, we perform a comprehensive ablation study. As summarized in Table~\ref{tab:ablation}, we systematically investigate the impact of input modality, network depth by varying the number of DSSC Block, key model hyperparameters including kernel size, pooling stride and the number of filters, as well as the convolution type by comparing depthwise separable approach against standard 1D convolutions. Each variant is evaluated in terms of overall accuracy (ACC), model size (Params), and computational complexity (FLOPs) to analyze the performance-efficiency trade-off.

The results demonstrate that the proposed model configuration achieves the best balance between performance and efficiency. For instance, Using EEG/EOG/EMG inputs with channel-wise processing further improves ACC (86.88\%) over using only EEG (85.58\%), indicating that modality-specific features are effectively captured. Besides, expanding the number of filters up to 128 channels yields the highest ACC (87.16\%), yet this 0.3\% improvement comes at an 8-fold increase in parameter count (from 13.3K to 101.5K). Furthermore, replacing standard convolutions with depthwise separable convolutions improves both accuracy (from 86.32\% to 86.88\%) and efficiency, reducing computational cost from 15.03M to 7.89M FLOPs. The optimal structure consists of three DSSC Blocks with kernel size 3, pooling stride 2, and filter configuration (8, 16, 32).

\subsection{Comparison with SOTA methods}

To validate the effectiveness and efficiency of our proposed ULW-SleepNet, we compare its performance against several state-of-the-art (SOTA) sleep stage scoring methods. Comparative results on the Sleep-EDF-20 and Sleep-EDF-78 datasets are presented in Table ~\ref{tab:results}. Performance is assessed using ACC, macro F1-score (MF1), Cohen’s kappa coefficient ($\kappa$), and per-class F1 scores, along with Params and FLOPs.


As shown in Table ~\ref{tab:results}, ULW-SleepNet demonstrates superior performance on Sleep-EDF-20, achieveing ACC of 86.9\%, F1-score of 80.7\%, and Cohen's Kappa coefficient of 0.82, outperforming all SOTA methods while utilizing only 13.3K parameters. On Sleep-EDF-78, ULW-SleepNet also delivers highly competitive results (ACC of 81.4\% , and F1-score of 74.0\% ) with the best Wake stage classification (F1-score of 93.0\% ). Although the accuracy is slightly lower than that of TinySleepNet and LWSleepNet by 1.8\% and 0.1\%, respectively, ULW-SleepNet achieves a substantial reduction in model size, requiring only 13.3K parameters, which corresponds to a 98.6\% decrease compared to TinySleepNet (1.3M) and a 97.8\% reduction relative to LWSleepNet (180K). Furthermore, the computational cost is significantly reduced, with FLOPs decreasing from 55.3M (LWSleepNet) to 7.89M, representing an 85.7\% reduction. It is worth noting that PicoSleepNet attains notably lower FLOPs (0.68M–0.73M), likely attributable to its use of Spiking Neural Networks, for which FLOPs-based metrics are not directly comparable to conventional CNNs due to the event-driven computation paradigm and hardware-dependent sparsity. Exploring such architectures may be a promising direction for future work to further decrease computational complexity.


\section{Conclusion and future work}
\label{sec:dis}

This paper introduces ULW-SleepNet, an ultra-lightweight and efficient framework designed for multimodal sleep stage scoring, addressing the challenges of high computational cost and large model size in wearable and resource-constrained applications. ULW-SleepNet incorporates an innovative DSSC Block together with several lightweight design strategies, including depthwise separable convolutions, channel-wise parameter sharing, and global average pooling. These elements collectively achieve a substantial reduction in both parameter count and computational complexity without compromising feature extraction capabilities. Comprehensive evaluations on the public Sleep-EDF-20 and Sleep-EDF-78 datasets demonstrate that ULW-SleepNet achieves an exceptional balance between performance and efficiency. 

Future work will examine cross-session robustness to assess longitudinal generalization, evaluate ULW-SleepNet on larger and more diverse clinical cohorts, and deploy the model on embedded hardware platforms for practical validation in real-world sleep monitoring systems.

\section{Acknowledgment}
This work was partially supported by  the Natural Science Foundation of Liaoning Province (2025-BS-0079), the Fundamental Research Funds for Central Universities (DUT25YG226, DUT24YG130), the Science and Technology Planning Project of Liaoning Province (2021JH1/10400049), the Jenny and Antti Wihuri Foundation, and the Scholarship from China Scholarship Council (202406060033).

\bibliographystyle{IEEEbib}
\bibliography{refs}

\end{document}